\documentclass[11pt]{article}

\usepackage[final]{acl}

\usepackage{times}
\usepackage{latexsym}
\usepackage{amsmath}
\usepackage[T1]{fontenc}

\usepackage[utf8]{inputenc}

\usepackage{microtype}

\usepackage{inconsolata}

\usepackage{graphicx}

\usepackage{amssymb,amsfonts,amsthm}
\usepackage{mathtools}
\usepackage{hyperref}
\usepackage{cleveref}
\usepackage{booktabs}
\usepackage{xcolor}
\usepackage{enumitem}
\usepackage{bm}
\usepackage{algorithm}
\usepackage{algpseudocode}

\theoremstyle{remark}

\newcommand{\method}{HiLP}
\newcommand{\nextlat}{NextLat}

\newcommand{\mtp}{MTP}

\newcommand{\ntp}{NTP}

\newcommand{\Lcal}{\mathcal{L}}
\newcommand{\E}{\mathbb{E}}

\newcommand{\Pbb}{\mathbb{P}}
\newcommand{\hh}{\mathbf{h}}
\newcommand{\uu}{\mathbf{u}}
\newcommand{\sg}{\mathrm{sg}}

\title{Hierarchical Latent Prediction for Language
Models}

\author{
 \textbf{Chang Shi\textsuperscript{1}\thanks{\ Work done during internship at Microsoft Research. Correspondence to: Chang Shi \href{changshi@utexas.edu}{<changshi@utexas.edu>}}},
 \textbf{Tim Pearce\textsuperscript{2}},
 \textbf{Manan Tomar\textsuperscript{2}},
 \textbf{Siddhartha Sen\textsuperscript{2}},
 \textbf{John Langford\textsuperscript{2}}
\\
 \textsuperscript{1}University of Texas at Austin,
 \textsuperscript{2}Microsoft Research
}

\begin{document}
\maketitle
\begin{abstract}
While standard Next-Token Prediction (NTP) lays the foundation of language model pretraining, its teacher-forced training paradigm may not be optimal for long-horizon reasoning and planning. Recent works such as Multi-Token Prediction (MTP) and Next-Latent prediction (NextLat) try to mitigate the problem through predicting multiple future tokens and self-supervised prediction in the latent space. However, those auxiliary objectives either have a limited horizon or suffer from compounding error from multi-step rollout. We introduce \textbf{\underline{Hi}erarchical \underline{L}atent \underline{P}rediction} (\method{}), which introduces an auxiliary higher-level abstract latent to help reduce the error accumulation effect in latent-space rollouts. Experiments show that \method{} can lead to longer-horizon coherent belief state representation and demonstrate the effectiveness of our method across coding and multi-step reasoning benchmarks, and offers more speculative decoding efficiency.
\end{abstract}

\section{Introduction}
\label{sec:introduction}

Next-token prediction (\ntp{}) with teacher forcing has yielded transformers of remarkable capability, yet it also introduces a training-inference mismatch known as exposure bias: during inference time, the auto-regressive token generation relies on its own previous outputs, leading to compounding errors which could degrade long-range generation quality. 

Recent work \nextlat{} \citep{teoh2025next} partially mitigates the issue through enforcing a latent space self-prediction loss. However, its latent dynamics model predicts one step ahead. This means that the self-predictive learning signal, while effective in inducing local transition consistency, provides only indirect gradient pressure to form representations that capture structure at longer temporal scales. In practice, long-horizon dependencies still remain attenuated by sequential dynamics unrolling.

Multi-scale temporal abstraction methods have a long history in sequence modeling and reinforcement learning \citep{sutton1999between,bacon2017option}. Inherently, language incorporates hierarchical structures at multiple scales, from characters, phrases to sentences, paragraphs and discourses. Therefore, it is reasonable to extend \nextlat{} with hierarchical learning signals at different granularities, for the sake of richer latent transition dynamics. This observation motivates a natural question: \textit{can we introduce temporal hierarchy directly into the latent-space self-prediction objective, shaping the representation of a language model transformer to encode multi-scale predictive structure during pretraining?}

In this paper, we answer affirmatively by introducing \textbf{Hierarchical Latent Prediction (\method{})}. Our contributions are as follows:
\begin{enumerate}[leftmargin=2em]
  \item We propose \method{}, a hierarchical representation training method that introduces multi-scale self-predictive learning into transformer pretraining via sliding-window attention over latents, a higher-level dynamics model, and a combined \ntp{} head.
  
  \item We evaluate the resulting models on downstream benchmark accuracy and speculative decoding efficiency, comparing \method{} with \mtp{} and \nextlat{}.
  
  \item We demonstrate that the entire hierarchical apparatus can be removed at inference time with no architectural overhead, preserving \nextlat{}'s option of being a pure training-time intervention.
\end{enumerate}

\section{Methodology}
We use $X_{1:T}$ to denote the token sequence prefix. A transformer $G_\theta$ produces
\emph{lower-level} latents $\hh_t = G_\theta(X_{1:t})$.
A sliding-window attention (SWA) module with window $k$ produces \emph{higher-level} abstract latents
$\uu_t = \mathrm{SWA}(\hh_{t-k+1:t})$,
a deterministic function of the last $k$ lower states.

\begin{figure}[ht]
  \includegraphics[width=\columnwidth]{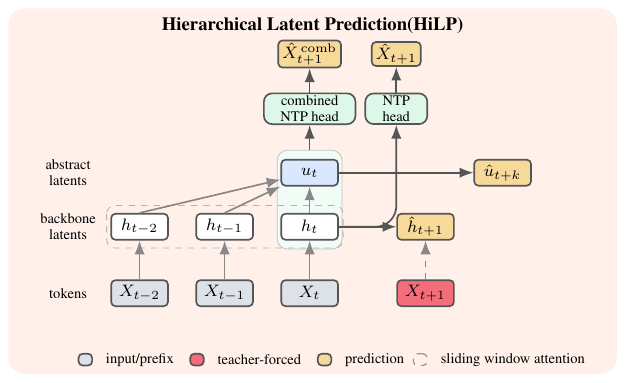}
  \caption{Overview of \method{} with sliding-window size $W=3$. The higher-level latent is computed by sliding-window attention over backbone latents, the higher next-latent predictor maps $u_t$ to $\hat{u}_{t+W}$ without token input, and both the standard \ntp{} head and combined \ntp{} head are trained, but the combined NTP head is dropped during inference to ensure no slow down in inference speed.}
  \label{fig:method}
\end{figure}

A higher-level dynamics model is trained to predict $k$ steps ahead in this abstract space. This explicit coarser-scale prediction objective encourages lookahead planning beyond single-step latent transitions. A combined \ntp{} head then conditions the next-token prediction on both level latents, taking advantage of abstract lookahead information to improve subsequent predictions. Four parameterised maps are jointly learned:
output head $p_\theta$, lower dynamics $p_\psi$,
upper dynamics $p_\phi$, and combined head $p_\rho$. An illustration of the architecture is shown in Fig. \ref{fig:method}.

\paragraph{Training objectives.}
\noindent
The total training objective is then a weighted sum of five terms: the standard \ntp{} loss, the \nextlat{} (lower-level) transition consistency loss, the KL term, the higher-level transition consistency loss, and the combined \ntp{} loss. At inference time, only the standard \ntp{} head is used, the entire hierarchical apparatus serves purely as an auxiliary training signal.

\begin{align}
\mathcal{L}_{\mathrm{ntp}}
  &= \mathbb{E}_{t<T}\!\left[
     -\log p_\theta(X_{t+1}\mid\hh_t)\right]
  \label{eq:A}\tag{A}\\
\mathcal{L}_{\mathrm{h}}
  &= \mathbb{E}_t\!\left[\tfrac{1}{d}
     \textstyle\sum_{i=1}^{d}
     \mathrm{SL1}(\sg[\hh_{t+i}], \hat{\hh}_{t+i})\right]
  \label{eq:B}\tag{B}\\
\mathcal{L}_{\mathrm{KL}}
  &= \mathbb{E}_t\!\left[\tfrac{1}{d}
     \textstyle\sum_{i=1}^{d}
     D_{\mathrm{KL}}\!\left(
       p_\theta^{\sg}(\cdot\mid\sg[\hh_{t+i}])
     \,\middle\|\right.\right. \notag\\
  &\quad\quad \left.\left.
       p_\theta^{\sg}(\cdot\mid\hat{\hh}_{t+i})
     \right)\right]
  \label{eq:C}\tag{C}\\
\mathcal{L}_{\mathrm{u}}
  &= \mathbb{E}_t\!\left[
     \mathrm{SmoothL1}(\sg[\uu_{t+k}],\,\hat{\uu}_{t+k})
     \right]
  \label{eq:E}\tag{D}\\
\mathcal{L}_{\mathrm{cntp}}
  &= \mathbb{E}_{t<T}\!\bigl[ \notag\\
  &\quad -\log p_\rho(X_{t+1}\mid\hh_t,\tilde{\uu}_t)
     \bigr]
  \label{eq:H}\tag{E}
\end{align}

Where $\mathrm{SL1}(\cdot,\cdot)$ denotes the SmoothL1 loss, $\sg[\cdot]$ the stop-gradient operator, and
$\tilde{\uu}_t \!:=\! \mathrm{SWA}(\sg[\hh_{t-k+1:t}])$
the higher-level latent with $\sg$ applied inside SWA
so gradients do not flow back into $\hh$.
We use this stop-gradient in the combined \ntp{} path to keep the token-level representation $\hh_t$ governed by the standard \ntp{} and lower-level transition consistency losses, while allowing the combined head to train
the SWA module and its higher-level representation. Without this separation, the combined-head cross-entropy would also update the lower latents through the SWA window, double-counting token-level supervision and creating a competing optimization signal for $\hh_t$.

\paragraph{Consistency conditions.}
At optimality, the losses enforce:
\begin{align}
p_\theta(X_{t+1}\mid\hh_t)
  &= \Pbb(X_{t+1}\mid X_{1:t})
  \tag{I}\label{eq:I}\\
p_\psi(\hh_{t+1}\mid\hh_t,X_{t+1})
  &= \Pbb(\hh_{t+1}\mid X_{1:t+1})
  \tag{II}\label{eq:II}\\
p_\phi(\uu_{t+k}\mid\uu_t)
  &= \Pbb(\uu_{t+k}\mid X_{1:t})
  \tag{III}\label{eq:III}\\
p_\rho(X_{t+1}\mid\hh_t,\uu_t)
  &= \Pbb(X_{t+1}\mid X_{1:t})
  \tag{IV}\label{eq:IV}
\end{align}
where (I) and (IV) enforce next-token consistency, (II) and (III) enforce transition consistency.

\section{Experiments}
1B-parameter models are trained on 100B tokens using 8 $\times$ NVIDIA B200 GPUs. 

\subsection{Coding and Multi-step Reasoning Benchmarks}
\begin{figure*}[ht]
  \includegraphics[width=1.8\columnwidth]{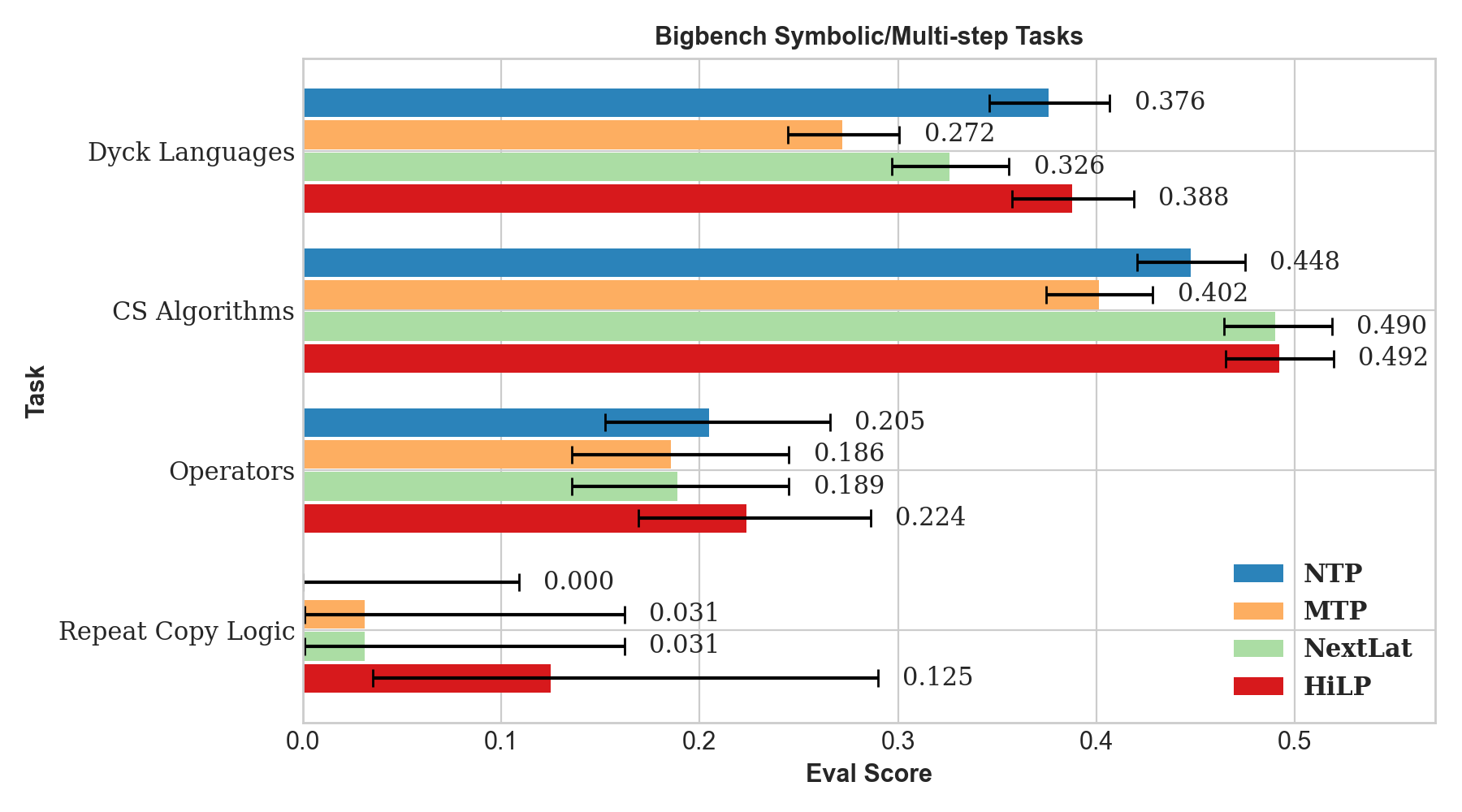}
  \caption{DCLM symbolic/multi-step eval results}
  \label{fig:dclm-symbolic-multistep}
\end{figure*}

After pretraining, we use LM Evaluation Harness \citep{eval-harness} to evaluate the zero-shot performance of the models on HumanEval coding benchmark \cite{chen2021codex}, and DataComp for LLMs \citep{li2025datacomplmsearchgenerationtraining}
to evaluate the models on a set of symbolic and multi-step benchmarks. Results in Tab. \ref{tab:benchmark-accuracy} and Fig. \ref{fig:dclm-symbolic-multistep} show that HiLP is improve performance from the multi-scale latent prediction.

\begin{table}[t]
  \centering
  \small
  \begin{tabular}{lc}
    \hline
    \textbf{Model} & \textbf{HumanEval pass@1} \\
    \hline
    \ntp{}     &  8.77 \\
    \mtp{}     &  9.21 \\
    \nextlat{} &  10.58 \\
    \method{}  &  11.33 \\
    \hline
  \end{tabular}
  \caption{Code-generation benchmark accuracy on HumanEval. HumanEval results are computed from 1000 samples per task.}
  \label{tab:benchmark-accuracy}
\end{table}

\begin{figure*}[th]
  \centering
  \includegraphics[width=\textwidth]{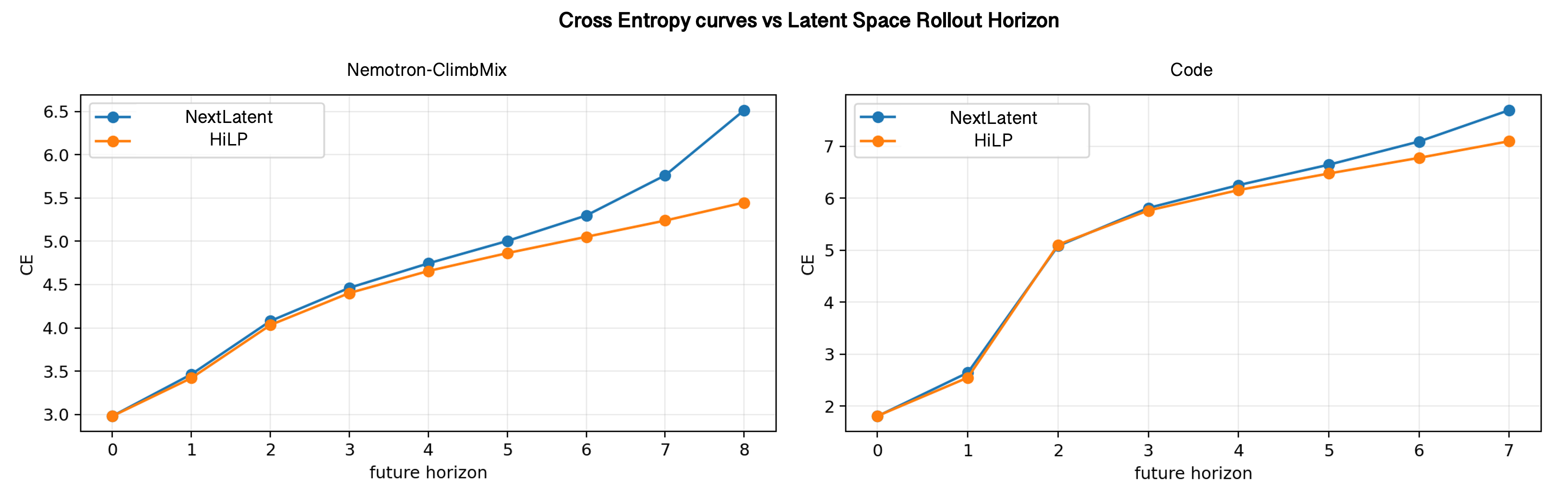}
  \caption{Latent cross-entropy curves for \nextlat{} and \method{} on web and code splits. \method{} has comparable near-term latent prediction loss and lower long-horizon future and excess cross-entropy, indicating more stable multi-step latent rollout.}
  \label{fig:latent-ce-curves}
\end{figure*}

\subsection{Speculative Decoding}
We evaluate speculative decoding on held-out validation splits of code and Nemotron-Climbmix data\citep{diao2026nemotron}. For each model, we report the \textbf{average number of accepted tokens} per drafting step, and \textbf{Avg match}, the per-position draft-verifier argmax agreement averaged over $K{=}1{\ldots}4$. Note that $K{=}0$ NTP position is always accepted and is omitted from the per-$K$ columns;


\begin{table*}[t]
  \centering
  \small
  \setlength{\tabcolsep}{4pt}
  \begin{tabular}{l|cc|cccccccc}
    \hline
    \textbf{Model} & \textbf{Avg Acc. Tok.}$\uparrow$ & \textbf{Avg match}$\uparrow$ & $K{=}1$ & $K{=}2$ & $K{=}3$ & $K{=}4$ & $K{=}5$ & $K{=}6$ & $K{=}7$ & $K{=}8$ \\
    \hline
    \mtp{}                                    & 2.57 & 0.545 & 0.707 & 0.633 & 0.531 & 0.310 & -- & -- & -- & -- \\
    \nextlat{}                                & 3.26 & 0.580 & 0.918 & 0.652 & 0.422 & 0.328 & 0.309 & 0.285 & 0.234 & 0.234 \\
    \method{} (NTP head)                      & 3.41 & 0.592 & 0.938 & 0.645 & 0.414 & 0.367 & 0.324 & 0.324 & 0.293 & 0.340 \\
    \hline
  \end{tabular}
  \caption{Speculative decoding results evaluated on a held-out validation split of code data.}
  \label{tab:speculative-decoding}
\end{table*}

\begin{table*}[t]
  \centering
  \small
  \setlength{\tabcolsep}{4pt}
  \begin{tabular}{l|cc|cccccccc}
    \hline
    \textbf{Model} & \textbf{Avg Acc. Tok.}$\uparrow$ & \textbf{Avg match}$\uparrow$ & $K{=}1$ & $K{=}2$ & $K{=}3$ & $K{=}4$ & $K{=}5$ & $K{=}6$ & $K{=}7$ & $K{=}8$ \\
    \hline
    \mtp{}                                    & 2.11 & 0.162 & 0.301 & 0.141 & 0.109 & 0.098 & -- & -- & -- & -- \\
    \nextlat{}                                & 2.57 & 0.483 & 0.641 & 0.504 & 0.410 & 0.375 & 0.324 & 0.312 & 0.246 & 0.254 \\
    \method{} (NTP head)                      & 2.59 & 0.491 & 0.699 & 0.492 & 0.379 & 0.395 & 0.328 & 0.406 & 0.289 & 0.289 \\
    \hline
  \end{tabular}
  \caption{Speculative decoding results evaluated on a held-out validation split of Nemotron-ClimbMix data.}
  \label{tab:speculative-decoding-climbmix}
\end{table*}

\section{Discussion}

\paragraph{Longer horizon prediction.}
The latent prediction cross-entropy losses across rollout horizons up to 8 steps ahead are shown in \ref{fig:latent-ce-curves}. As the curves show, \method{} preserves the near-term cross-entropy of \nextlat{} while producing lower future prediction error at longer horizons, suggesting that the hierarchical latent mitigates multi-step error accumulation.

\section{Related Work}
\label{sec:related}

Our work sits at the intersection of several active research threads: pretraining objectives beyond next-token prediction, latent-space language modeling, and hierarchical models.

\subsection{Pretraining Objectives Beyond Next-Token Prediction}
\label{sec:related:beyond_ntp}
The next-token prediction objective has largely limited the capacity of language models in downstream tasks that require longer-horizon reasoning and planning\citep{bachmann2024pitfalls, nagarajan2025rolldicelook}. Recent works have started to introduce auxiliary learning signals to mitigate this myopic gap through further future prediction \citep{gloeckle2404better, ahn2025efficient, teoh2025next, mahajan2025beyond}. However, these approaches usually operate on a single granularity scale and have limited horizon capability due to compounding error. \method{} differs in two ways: it introduces two latent prediction pathways at different temporal scales, so the coarser pathway supervises long-horizon structure directly rather than through repeated single-step unrolling; and all auxiliary heads are dropped at inference, so the longer-horizon signal costs nothing at deployment time.

\subsection{Latent-Space Language Modeling}
\label{sec:related:latentgen}
Another line of work moves language generation itself into a continuous latent space. Large Concept Models \citep{barrault2024large} perform autoregressive prediction over sentence-level embeddings, treating each sentence as a ``concept'' in a shared representation space. CALM \citep{shao2025continuous} replaces next-token prediction with next-vector prediction, compressing a chunk of $K$ tokens into a single continuous vector to raise the semantic bandwidth of each generative step. Coconut \cite{shao2025continuous} lets the model reason in latent space by feeding its own hidden state back as the next input embedding instead of decoding to tokens. All of these approaches change the inference-time generation process to operate on latents, requiring bespoke decoding procedures or latent-to-text decoders. \method{} is the opposite design point: latent prediction is used purely as an auxiliary training signal to shape representations, while inference remains standard token-level autoregressive decoding with zero added latency.

\subsection{Hierarchical Models}
\label{sec:related:hierarch}
Hierarchical model design has been researched in both language modeling area and other domains.

MegaByte \citep{yu2023megabyte} stacks a global patch-level transformer over a local byte-level one, the Byte Latent Transformer \citep{pagnoni2025byte} dynamically segments bytes into entropy-based patches that serve as the units of computation, and H-Net \cite{hwang2025dynamic} learns content-dependent chunking end-to-end within a hierarchical U-Net-like network. 

In model-based reinforcement learning, Hierarchical Planning with Latent World Models \citep{zhang2026hierarchical} learns world models at multiple temporal scales within a shared latent space, using long-horizon latent predictions as subgoals for short-horizon control. In representation learning theory, Learning Discrete Concepts in Latent Hierarchical Models \citep{kong2024learning} formalizes concepts as discrete latent variables organized in a hierarchical causal model and derives identifiability conditions for recovering such hierarchical concept structure from high-dimensional unsupervised data. In time series analysis. HiTime \cite{tao2024hierarchical} employs a hierarchical feature encoder together with a hybrid prompting strategy to align time series and text modalities, improving multivariate time series classification with large language models. These works demonstrate the benefit of hierarchical latent structure across planning, representation identifiability, and time series domains. \method{} differs in a way that the hierarchy complexity is not consumed at inference time: to our knowledge it is the first to introduce hierarchical latent prediction as a pretraining objective for autoregressive language models, where the hierarchy shapes the representation during training and is then removed entirely.

\section{Conclusion}
HiLP introduces temporal hierarchy into the latent-space self-prediction objective, shaping the representation of language model transformers to encode multi-scale predictive structure during pretraining. Experiments show that it improves both speculative decoding and language models on tasks that need longer-horizon reasoning and planning.

\section{Future Work}
In current work, the abstract latent prediction horizon is a manually set hyperparameter,  which limits the flexibility of lookahead planning. Further designs including dynamically choosing the lookahead horizon could be beneficial. Also, the NTP head is used during inference in the current implementation, but combined NTP can also be used if we trade speed for accuracy. 





\bibliography{custom}


\newpage
\appendix
\section{HiLP Training Procedure}
\label{sec:appendix-algorithm}

Algorithm~\ref{alg:hilp-training-step} summarizes one \method{} training step, combining the five objectives into a single backward pass. The SWA module is a causal sliding-window self-attention over the lower latents, so every position $t$ carries its own higher-level latent $\uu_t$ summarizing the last $k$ lower latents. All latent-space prediction targets are stop-gradiented, and both distributions in the KL term are decoded through a frozen copy of the \ntp{} head ($p_\theta^{\sg}$), so each auxiliary loss shapes the representation only through its designated pathway: the rollout losses reach $\hh$ through $p_\psi$, the higher-level consistency loss trains $p_\phi$ and the SWA module, and the combined cross-entropy trains $p_\rho$ and the SWA module alone via the stop-gradient inside $\tilde{\uu}_t$. At inference, only $G_\theta$ and $p_\theta$ are kept: the SWA module, both dynamics models, and the combined head are discarded.

\begin{algorithm*}[t]
  \small
  \caption{One \method{} training step. $E(\cdot)$ denotes the backbone, $k$ the SWA window size and lookahead offset, and $d$ the rollout depth. Expectations over $t$ average over all valid positions in the batch.}
  \label{alg:hilp-training-step}
  \begin{algorithmic}[1]
    \Require batch $X_{1:T}$; backbone $G_\theta$ with \ntp{} head $p_\theta$; lower dynamics $p_\psi$; SWA module; higher dynamics $p_\phi$; combined head $p_\rho$; loss weights $\lambda_{\mathrm{ntp}}, \lambda_{\mathrm{h}}, \lambda_{\mathrm{KL}}, \lambda_{\mathrm{u}}, \lambda_{\mathrm{cntp}}$
    \State $\hh_{1:T} \gets G_\theta(X_{1:T})$ \Comment{one backbone pass}
    \State $\Lcal_{\mathrm{ntp}} \gets \E_t\big[{-\log p_\theta(X_{t+1}\mid\hh_t)}\big]$ \Comment{Eq.~\eqref{eq:A}}
    \Statex \textit{Lower-level rollout (teacher-forced tokens):}
    \State $\hat\hh_t \gets \hh_t$ for all $t$; \quad $\Lcal_{\mathrm{h}}, \Lcal_{\mathrm{KL}} \gets 0$
    \For{$i = 1, \dots, d$}
      \State $\hat\hh_{t+i} \gets p_\psi\big(\hat\hh_{t+i-1},\, E(X_{t+i})\big)$
      \State $\Lcal_{\mathrm{h}} \mathrel{+}= \tfrac{1}{d}\,\E_t\big[\mathrm{SL1}\big(\sg[\hh_{t+i}],\, \hat\hh_{t+i}\big)\big]$ \Comment{Eq.~\eqref{eq:B}}
      \State $\Lcal_{\mathrm{KL}} \mathrel{+}= \tfrac{1}{d}\,\E_t\big[D_{\mathrm{KL}}\big(p_\theta^{\sg}(\cdot\mid\sg[\hh_{t+i}]) \,\big\|\, p_\theta^{\sg}(\cdot\mid\hat\hh_{t+i})\big)\big]$ \Comment{Eq.~\eqref{eq:C}}
    \EndFor
    \Statex \textit{Higher-level channel:}
    \State $\uu_t \gets \mathrm{SWA}(\hh_{t-k+1:t})$ \Comment{abstract latent}
    \State $\hat\uu_{t+k} \gets p_\phi(\uu_t)$ \Comment{$k$-step lookahead, no token input}
    \State $\Lcal_{\mathrm{u}} \gets \E_t\big[\mathrm{SmoothL1}\big(\sg[\uu_{t+k}],\, \hat\uu_{t+k}\big)\big]$ \Comment{Eq.~\eqref{eq:E}}
    \Statex \textit{Combined head (sg keeps $\hh$ governed by \ntp{} and rollout):}
    \State $\tilde\uu_t \gets \mathrm{SWA}\big(\sg[\hh_{t-k+1:t}]\big)$
    \State $\Lcal_{\mathrm{cntp}} \gets \E_t\big[{-\log p_\rho(X_{t+1}\mid\hh_t,\tilde\uu_t)}\big]$ \Comment{Eq.~\eqref{eq:H}}
    \State $\Lcal \gets \lambda_{\mathrm{ntp}}\Lcal_{\mathrm{ntp}} + \lambda_{\mathrm{h}}\Lcal_{\mathrm{h}} + \lambda_{\mathrm{KL}}\Lcal_{\mathrm{KL}} + \lambda_{\mathrm{u}}\Lcal_{\mathrm{u}} + \lambda_{\mathrm{cntp}}\Lcal_{\mathrm{cntp}}$
    \State One optimizer step on $\nabla\Lcal$ w.r.t.\ $\theta$, $\psi$, $\phi$, $\rho$, and the SWA module
  \end{algorithmic}
\end{algorithm*}

\section{Experiment hyperparameters}
\begin{table*}[t]
  \centering
  \small
  \begin{tabular}{p{0.18\linewidth}p{0.46\linewidth}p{0.24\linewidth}}
    \toprule
    \textbf{Group} & \textbf{Hyperparameter} & \textbf{Value} \\
    \midrule
    Backbone model & Vocabulary size & 100{,}352 \\
                   & Training sequence length & 8{,}192 \\
                   & Hidden layers   & 24 \\
                   & Attention heads & 16 \\
    \midrule
                 & Latent input combination & GLU-cross \\
    Architecture & Hidden multiplier & 2 \\
                 & Latent window size & 4 \\
                 & Latent lookahead steps & 4 \\
    \midrule
                 & NTP loss weight & 1.0 \\
                 & MSE loss weight & 10.0 \\
    Loss weights & KL loss weight & 0.1 \\
                 & Higher-level MSE loss weight & 1.0 \\
                 & Combined NTP loss weight & 0.5 \\
    \midrule
                  & Input combination & GLU-cross \\
    Combined head & Downstream LM head & NTP \\
                  & Draft LM head & NTP \\
    \midrule
    Optimization & Latent learning rate & $1.0 \times 10^{-3}$ \\
    Latent evaluation & Evaluation horizon & 8 \\
    \bottomrule
  \end{tabular}
  \caption{HiLP hyperparameter configuration.}
  \label{tab:latent-parameters}
\end{table*}

\newpage
\section{Latent overhead and efficiency.}
Table~\ref{tab:efficiency-comparison} compares the four 100B-token code-data runs along three axes: total parameter count, the subset of
parameters actually used during greedy NTP decoding, and training
throughput. All four models share the same 1.06\,B-parameter trunk
and tied-free LM head, so the verifier path used at inference is
identical and the additional parameters in \mtp{}, \nextlat{}, and
\method{} are auxiliary draft components that are not required to
produce the next-token distribution. \nextlat{} adds a small
($\approx$19\,M) next-latent predictor on top of the trunk, while
\method{} and \mtp{} each add a $\approx$190--200\,M draft module.
Training throughput scales inversely with the size of the auxiliary
loss graph: the NTP baseline reaches $\approx$126\,K tokens/s/GPU,
\nextlat{} retains $\approx$83\% of that throughput, \method{}
retains $\approx$65\%, and \mtp{} drops to $\approx$49\% because its
four future-token heads must each be evaluated against the LM logits
at every step.

\begin{table*}[ht]
  \centering
  \small
  \begin{tabular}{lrr rrrr rrrr}
    \hline
    \textbf{Model}
      & \textbf{\# Params} & \textbf{Inf. Params}
      & \multicolumn{4}{c}{\textbf{Training Step Breakdown (ms)}}
      & \multicolumn{4}{c}{\textbf{Training Throughput}} \\
    \cline{4-7}\cline{8-11}
      & & & \textbf{Step} & \textbf{Fwd} & \textbf{Bwd} & \textbf{Optim} 
      & \textbf{Tok/s/GPU}$\uparrow$ & \textbf{TF/GPU}$\uparrow$ & \textbf{MFU}$\uparrow$ & \textbf{Samp/s}$\uparrow$ \\
    \hline
    \ntp{}     & 1.06\,B & 1.06\,B & 449 & 119 & 293 & 37 & 126{,}278 & 898 & 0.399 & 123.3 \\
    \mtp{}     & 1.25\,B & 1.06\,B & 916 & 631 & 243 & 43 &  61{,}789 & 509 & 0.226 &  60.3 \\
    \nextlat{} & 1.08\,B & 1.06\,B & 542 & 205 & 295 & 42 & 104{,}505 & 884 & 0.393 & 102.1 \\
    \method{}  & 1.27\,B & 1.06\,B & 690 & 329 & 315 & 46 &  81{,}653 & 780 & 0.347 &  79.7 \\
    \hline
  \end{tabular}
  \caption{Parameter counts and training efficiency for the
    100B-token runs on $8\!\times\!$B200 GPUs.
    \textbf{\# Params}: total trained parameters.
    \textbf{Inf.\ Params}: verifier path (trunk + LM head) used during greedy NTP decoding, identical across all models.
    \textbf{Step breakdown}: \texttt{timing/step\_time\_sec} and its
    components (forward, backward, optimizer), in
    milliseconds.
    \textbf{Throughput}: tokens/s/GPU, TFLOP/s/GPU, model FLOPs utilization, and samples/s (all GPUs).}
  \label{tab:efficiency-comparison}
\end{table*}

\newpage
\section{Input-Combination Modes}

We include an implementation-level ablation comparing the historical
\texttt{concat} input-combination mode against the current
\texttt{glu\_cross} mode used by latent prediction modules.
Let $D$ denote the hidden size, $m$ the latent-head hidden multiplier, and $V$ the vocabulary size. In the lower next-latent predictor, \texttt{concat} forms $[\hh_t; e_{t+1}] \in \mathbb{R}^{2D}$ and feeds it directly to the SwiGLU predictor. This gives an intermediate width of $2mD$ and costs $10mD^2$ parameters and multiply-adds per token
for the gate, up, and down projections, ignoring biases. By contrast, \texttt{glu\_cross} computes
\[
  W_h \hh_t \odot \sigma(W_e e_{t+1}) \in \mathbb{R}^{D},
\]
adding two $D \times D$ projections but reducing the following SwiGLU
width to $mD$. Its corresponding cost is therefore
$(3m+2)D^2$. With the default $m=4$, this is $14D^2$ versus
$40D^2$, or about $35\%$ of the \texttt{concat} predictor cost.

The same distinction appears in the combined \ntp{} head. For
\texttt{concat}, the combined head maps $[\hh_t;\uu_t]\in
\mathbb{R}^{2D}$ to logits, costing $2DV$. For \texttt{glu\_cross},
the gated fusion costs $2D^2$ and the logits are produced from a
$D$-dimensional vector, costing $DV$, for a total of $DV+2D^2$.
Thus \texttt{glu\_cross} is especially attractive when $V \gg D$:
it approximately halves the combined-head logit computation while
also reducing the latent predictor from a $2D$-wide to a $D$-wide
SwiGLU input.

\begin{table}[ht]
  \centering
  \small
  \begin{tabular}{lcc}
    \hline
    \textbf{Mode} & \textbf{NTP loss}$\downarrow$ & \textbf{Step (ms)}$\downarrow$ \\
    \hline
    \texttt{concat}     & 4.113 & 919.5 \\
    \texttt{glu\_cross} & 3.942 & 818.9 \\
    \hline
  \end{tabular}
  \caption{Input-combination ablation comparing next-token prediction loss and training step time.}
  \label{tab:combination-mode-ablation}
\end{table}

\begin{figure}[ht]
\centering
  \includegraphics[width=0.9\columnwidth]{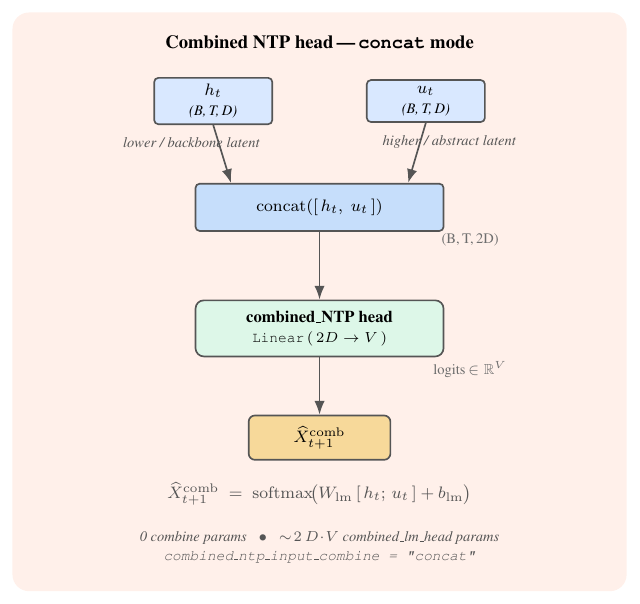}
  \caption{\texttt{concat} input-combination mode used by the latent and combined \ntp{} heads.}
  \label{fig:combination-concat}
\end{figure}

\begin{figure}[ht]
  \centering
  \includegraphics[width=0.9\columnwidth]{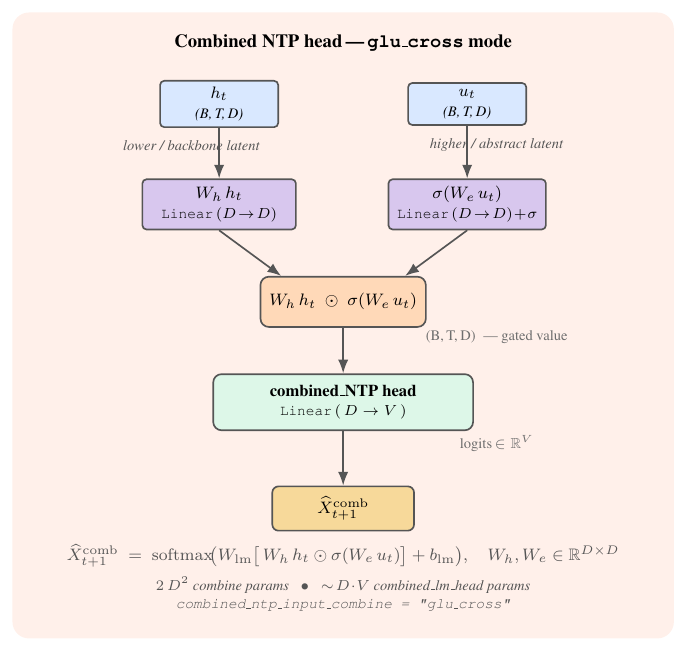}
  \caption{\texttt{glu\_cross} input-combination mode, which gates a projected lower latent with the conditioning latent while preserving hidden width $D$.}
  \label{fig:combination-glu-cross}
\end{figure}



\end{document}